\documentclass{cta-author}

{}
{}
{}
\usepackage{times}
\usepackage{epsfig}
\usepackage{graphicx}
\usepackage{amsmath}
\usepackage{amssymb}
\usepackage{indentfirst}
\usepackage{multirow}
\usepackage{stfloats}
\usepackage{enumerate}
\usepackage{caption}
\usepackage{subcaption}

\begin{document}

\supertitle{Research Article}

\title{Attentive pooling for Group Activity Recognition}

\author{\au{Ding Li$^{1,2}$}, \au{Yuan Xie$^{1,2}$}, \au{Wensheng Zhang$^{1,2}$}, \au{Yongqiang Tang$^{1,2}$}, \au{Zhizhong Zhang$^{1,2}$}} 

\address{\add{1}{Institute of Automation, Chinese Academy of Science, Beijing, 100190, People's Republic of China \\
		 \add{2}{Research Center of Precision Sensing and Control, Institute of Automation, Chinese Academy of Sciences, Beijing, 100190, People's Republic of China}
}
\email{zhangwenshengia@hotmail.com}}

\begin{abstract}
\looseness=-1 In group activity recognition, hierarchical framework is widely adopted to represent the relationships between individuals and their corresponding group, and has achieved promising performance. However, the existing methods simply employed max/average pooling in this framework, which ignored the distinct contributions of different individuals to the group activity recognition. In this paper, we propose a new contextual pooling scheme, named attentive pooling, which enables the weighted information transition from individual actions to group activity. By utilizing the attention mechanism, the attentive pooling is intrinsically interpretable and able to embed member context into the existing hierarchical model. In order to verify the effectiveness of the proposed scheme, two specific attentive pooling methods, \textit{i.e.}, global attentive pooling (GAP) and hierarchical attentive pooling (HAP) are designed. GAP rewards the individuals that are significant to group activity, while HAP further considers the hierarchical division by introducing subgroup structure. The experimental results on the benchmark dataset demonstrate that our proposal is significantly superior beyond the baseline and is comparable to the state-of-the-art methods.

\end{abstract}

\maketitle

\section{Introduction}\label{sec1}

Understanding human actions in a given video sequence has stimulated much research interest in computer vision. During the past several years, massive efforts \citep{simonyan2014two, wang2016temporal, ji20133d, tran2015learning, yu2017fully, varol2018long} have been made for the individual action recognition, which merely focuses on the actions of a single person. When multiple individuals are in the scene, rather than classifying person-level actions in isolation, recognizing the activities performed by the group facilitates a lot of applications, \textit{e.g.} video surveillance, sport analytics and video retrieval. Recently, the group activity recognition has received considerable academic attention \citep{choi2012unified,ibrahim2016hierarchical,shu2017cern,li2017sbgar,bagautdinov2017social,wang2017recurrent,biswas2018structural}. Compared with individual action recognition, the main challenge of group activity recognition is modelling hierarchical relationships between individuals and group.

To overcome this difficulty, numerous researches \cite{ibrahim2016hierarchical, DBLP:journals/corr/IbrahimMDVM16, biswas2018structural} have focused on the hierarchical structure, which are on the basis of the unity of individuals. Considering that individual actions collectively define the group activities, these approaches first establish a separate model for recognition and analysis of individual actions and its dynamics. Then, the representations of group activities are extracted by summarizing the information of collected individual actions. Typically in several deep learning approaches, Recurrent Neural Networks, \textit{e.g.} LSTM networks \cite{LSTM} have been used to capture the dynamics of actions and activities, and max pooling or average pooling are applied to the aggregation of person-level representations.

Despite the promising performance, the max/average pooling adopted in \cite{ibrahim2016hierarchical, DBLP:journals/corr/IbrahimMDVM16, shu2017cern, biswas2018structural} ignored the distinguished contributions of different individuals to the group activity recognition. For instance, there is a clip labelled as "Left Winpoint" shown in Figure 1(a). According to the unity of players in this clip, we can conclude that the team in the left side of the court are gathering and celebrating their scoring, while their rivals lose in this round. The clip shown in Figure 1(b) is labelled as "Right Spike", where a person in the right side of the court performs the individual action "spiking"(marked by the yellow rectangle), and two players across the net are blocking (marked by red rectangle). Meanwhile, there are several players performing "Moving" action around their teammates, and persons labelled as "Standing" or "Waiting" are preparing for competing in the next round. In fact, we are able to recognize the group activity of the frame by those people who perform "spiking" and "blocking", because others make less contributions to the final group activity prediction. It is the individual differences that facilitate the group activity recognition, and the more diverse are the individuals, the more complicated contextual relationship of the group get. Previous pooling scheme is able to aggregate features effectively when the contextual relationship is simple (Figure 1(a)), while achieved limited success when the contextual relationship gets complicated (Figure 1(b)).

\begin{figure}[tp]
	\centering
	\begin{minipage}{8cm}
		{\includegraphics[width=7.8cm,height=3.5cm]{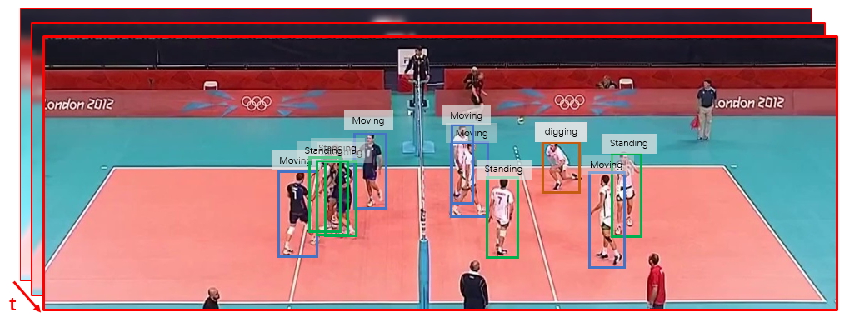}
			\subcaption{A video clip labeled as group activity "Left Winpoint"\label{subfig:above}}}
	\end{minipage}
	\begin{minipage}{8cm}
			{\includegraphics[width=7.8cm,height=3.5cm]{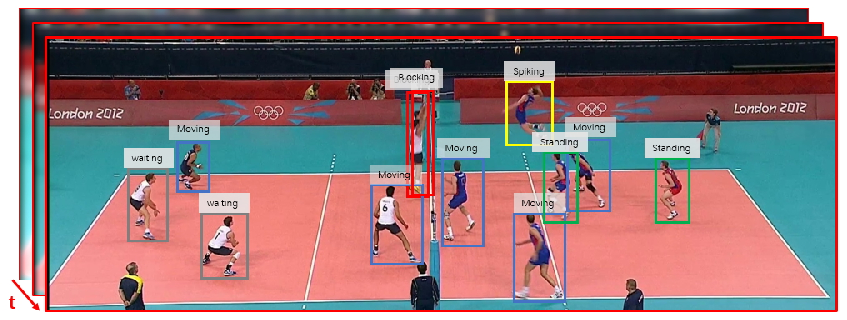}
			\subcaption{A video clip labeled as group activity "Right Spike"\label{subfig:bellow}}}
	\end{minipage}

	\caption{The illustration of our motivation in distinguishing individuals with different attention. Players with different actions in the scene are marked by bounding boxes with different colour.
		 \label{fig:{A frame labeled as group activity "Right Spike"}_{A frame labeled as group activity "Left Winpoint"}}}

\end{figure}

Thus, there is an imperative need to consider individual differences as well as the unity of the group simultaneously, and this motivates us to propose a novel pooling scheme, named attentive pooling. To verify the effectiveness of the proposed scheme, we first design an global attentive pooling (GAP) model, which rewards the individuals that are significant to group activity. In addition, it has been proven that the structure of the focal group is beneficial to analyse the group activity \cite{DBLP:journals/corr/IbrahimMDVM16}. Subgroups are introduced after the partition of the whole group. Hence, We further present a hierarchical attentive pooling (HAP) model, which attaches variable importance to both person-level and subgroup-level features.

Our main contributions of this paper can be summarized as follows: 
\begin{enumerate}[I. ]
	\item  We present a novel pooling scheme for group activity recognition, which can better explore the relationships between individual actions and group activities.
	\item  We extend the attentive pooling scheme to the hierarchical mode, and thoroughly compare the different modes of proposed scheme. 
	\item  The evaluation of attentive pooling scheme is conducted on the widely-used Volleyball dataset, and the results confirm the effectiveness of our proposal. 
\end{enumerate}

The rest of this paper is organized as follows. In Section 2, we briefly review some related works, followed by the introduction of several variants of attentive pooling based methods and implementation details in Section 3. Experimental results and discussions are showed in Section 4. Finally, we conclude this paper in Section 5.

\section{Related work}

Human activity recognition is a hot topic for research, surveys such as \citep{herath2017going,chaquet2013survey} have reviewed the vast literature in this area. Here, we will mainly focus on the group activity recognition and recent related advances in attention mechanism.

{\bfseries Group activity recognition.} In the early years, hand-crafted features are used as representation of individual and group-level activity in most previous approaches. Nabi \textit{et~al.} \citep{nabi2013temporal} proposed a pose-let activation pattern over time (TPOS) descriptor to capture human motions and interactions in groups. A unified framework \citep{choi2012unified} proposed by Choi \textit{et~al.} simultaneously achieve tracking multiple people, recognizing individual actions, interactions and collective activities in a joint framework. These methods which use hand-crafted features are highly dependent on hypothesis, and cannot be trained in an end-to-end fashion.

In recent years, deep learning has been widely used in group activity recognition, and shows great potential to accelerate the research process. Considering the hierarchical structure of group activity analysis in video, a framework \citep{ibrahim2016hierarchical} consists of multi-level cascade of recurrent neural networks was proposed, and it greatly inspired researchers in this area. The approach can be divided into two steps. First, tracklets of multi-person are constructed based on the detection and trajectories, and sptio-temporal features are extracted from these tracklets by utilizing deep convolutional neural network and lower recurrent neural network. Second, the extracted features of multi-person are fed into the higher recurrent neural network after passing through pooling module along the axis of persons. The final prediction of individual actions and group activities are obtained via softmax in a feed-forward way. \citep{DBLP:journals/corr/IbrahimMDVM16} extended the hierarchical framework by splitting the whole group into several subgroups, which improved the performance. Thereafter, several methods are proposed from different perspectives. Shu \textit{et~al.} \citep{shu2017cern} extended this hierarchical framework by specifying a novel energy layer and computing the corresponding p-values to estimate the most confident energy minimum. Bagautdinov \textit{et~al.} \citep{bagautdinov2017social} presented a unified framework for detecting and recognizing human social behaviors in raw image sequences. Li \textit{et~al.} \citep{li2017sbgar} proposed a semantic-based approach which generates captions from video frames and predict final activity categories based on generated captions. Biswas \textit{et~al.} \citep{biswas2018structural} proposed a structural recurrent neural network (SRNN) that uses a series of interconnected RNNs to jointly capture the actions of individuals, their interactions, as well as the group activity. These methods which employ max pooling operation have achieved clear improvement compared to those using handcraft features, but they ignore the distinct contributions of different individuals to the group activity recognition. To address this issue, Wang \textit{et~al.} \cite{wang2017recurrent} proposed a recurrent interactional context modelling scheme based on LSTM network and produces more discriminative/descriptive interactional features. However, this framework has three layers of LSTM network, which increases the complexity of the model.

{\bfseries Attention mechanism.} Attention mechanism is considered as content-based addressing when processing focal sequence, and has been applied to several tasks in computer vision and natural language processing \cite{vaswani2017attention}. According to the process of selecting attentive parts, attention mechanism can be roughly divided into two paths, named hard attention and soft attention. As a representative of the first path, Mnih \textit{et~al.} \citep{mnih2014recurrent} presented a recurrent neural network model that is capable of extracting information from an image or video by adaptively selecting a sequence of regions or locations and only processing the selected regions at high resolution. The hard attention block is trained as the training process of reinforcement learning, which may cause difficulties to converge. Instead of hard selections of parts of interest, soft attention mechanisms were proposed by using weighted averages. In the soft attention path, Sharma \textit{et~al.} \citep{sharma2015action} proposed a Soft-Attention LSTM model built on top of multi-layered RNNs to selectively focus on parts of the video frames and classify videos after taking a few glimpses. \cite{yang2016hierarchical} proposed a hierarchical attention network for document classification. In this approach, the network has two levels of attention mechanisms applied at the word and sentence-level, enabling it to attend differentially to more and less important content. Li \textit{et~al.} \cite{videolstm} presented an end-to-end sequence learning model for action recognition in video, which captures the spatial layout by hardwiring convolutions in the soft-attention LSTM. Long \textit{et~al.} \cite{attentionclusters2019} explored the potential of attention mechanism when aggregating local features, and proposed a attention clusters based framework for video classification.

Attention allows the model to avoid the negative impact of noisy parts, thus can improve performance of recognition. Specifically,  attention allows the model to assign a relevance score to the elements in the group, and highlights elements with the most task-relevant information. Further more, it reveals the structural information within the group, which provides a way for us to make results more interpretable.

\section{Method}

As mentioned before, the goal of this paper is to recognize activities of a group of people. We utilize a bottom-up approach to represent and recognize the individual actions and group activities in a hierarchical manner, and employ the attentive pooling scheme in the transition from person-level features to group-level features. Comprehensively understanding a single person action and its temporal dynamics is the foundation of group activity recognition, and modelling the relationships between individuals and their corresponding group in a hierarchical manner has been validated among researchers. By using attentive pooling scheme which is easy to be integrated into the hierarchical framework, we can take into account both individual differences and unity of the group.

\begin{figure*}[htb]
	\centering
	\includegraphics[width=14.8cm,height=7.8cm]{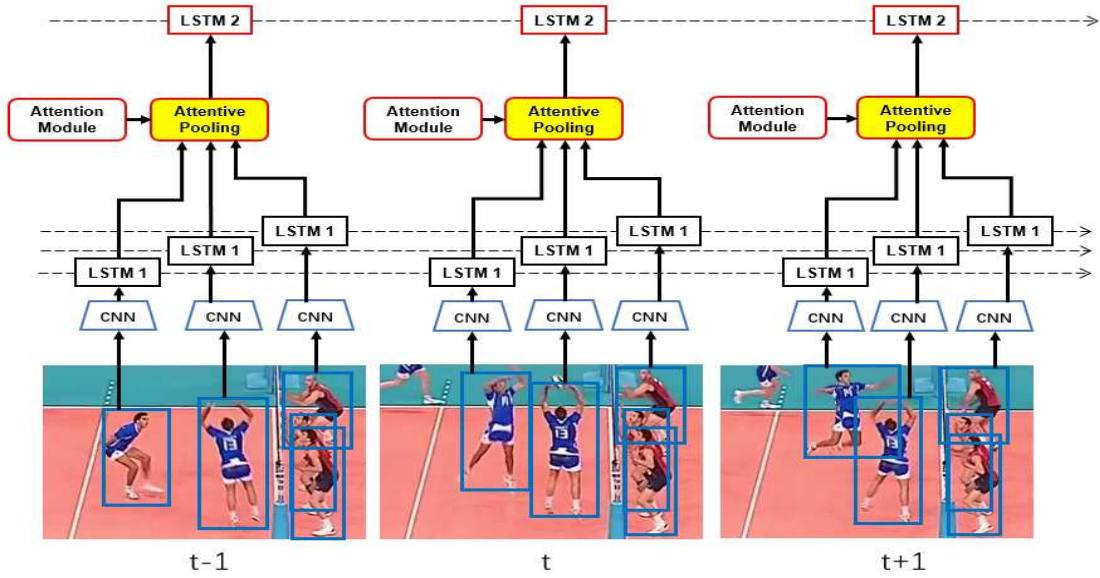}
	\caption{Attentive pooling based framework for group activity recognition. Taking a sequence of frames where N players compete as input, we extract the static and dynamic representations of each individual by CNN and low-level LSTM layers, denoted as black rectangles. After that, we feed the individual features into attentive pooling layer. The output of attentive pooling layer is passed through high-level LSTM layers and softmax layers trained to recognize group activities. The timeline is represented by the dotted arrow.}
	\label{fig2}
\end{figure*}

\subsection{Hierarchical Temporal Model(HTM) of Individual Action and Group Activity}
To clearly represent action information of a person, it is very necessary to capture the spatial features and its temporal dynamics simultaneously. As is common sense to us all, human action can be seen from both static and dynamic perspectives. There are several actions that can be discriminated from a single frame, while dynamic information extracted from multiple frames is needed for recognizing other actions. 

Inspired by the excellent performance of deep Convolutional Neural Network (CNN), we use CNN to extract features from the bounding box around the person in each time step on a person trajectory. Similar to \citep{donahue2015long}, we use long short-term memory (LSTM) models \citep{LSTM} to represent temporal dynamics of individual action. Such temporal information and spatial features can represent an individual action, and are critical for group performance. LSTM has been widely employed for many sequential problems in computer vision. The content of the memory cell in a LSTM unit is regulated by several gating units that control the flow of information in and out of the cells. The control they offer also helps in avoiding spurious gradient updates that can typically happen in training RNNs when the length of a temporal input is large. This property enables us to stack such layers in order to learn complex dynamics present in the input in different ranges. The output of the CNN, represented by $x_i$, can be considered as a region-based feature describing the appearance around a person.

For ${i^{th}}$ person in the scene, we denote the sequence of spatial features extracted by CNN by ${x_i} = \left( {{x_{i}^1},{x_{i}^2}, \ldots ,{x_{i}^T}} \right)$, and the sequence of temporal features ${h_i} = \left( {{h_{i}^1},{h_{i}^2}, \ldots ,{h_{i}^T}} \right)$. $T$ is the length of the sampled video clip. Then, $h_i$ will be fed into LSTM block where the forget gate, input gate and output gate are trained automatically, the update process at time-step t can be simply expressed as:
$${h_{i}^t} = LSTM\left( {{h_{i}^{t-1}},{x_{i}^t}} \right)$$
Besides, we choose to concatenate the spatial(static) and temporal (dynamic) feature (represented by $\oplus$), and get the comprehensive representation of single person's action, denoted as ${P_{i}}$. At time-step t, it can be expressed as follows:
$${P_{i}^t} = {h_{i}^t} \oplus {x_{i}^t}$$

As for a group of people, group-level activity is collectively defined by individual person-level actions. Once got spatio-temporal features of person-level action, the next step is to model the group-level activity in hierarchical architecture. The whole framework is shown as Fig. \ref{fig2}.

In order to get group-level activity information, we need to apply pooling operation to person-level spatio-temporal features. Here, we propose attentive pooling methods to aggregate person-level features. The aim of attentive pooling is to get a comprehensive collection of person-level information, so as to change our focal object into the whole group. Attentive pooling is the core of our method, and will be explicitly shown in next section. After that, we get the representation of the whole group, denoted as $G_t$. Finally, we can take group representation $G_t$ as the input of group-level LSTM, thus, we can get final recognition by analysing the hidden state of the second-stage LSTM layer. Explicitly, we make $G_t$ pass through a fully connected layer, then input the second-stage LSTM layer. The hidden state of the second-stage LSTM layer $H_{g}^t$ is the group-level features we want. $H_{g}^t$ carries temporal information for the whole group dynamics. In the end, we feed $H_{g}^t$ into a softmax classification layer to get final predictions for group activity. In order to train the network, we minimize the joint loss:
$$L = L\left( {{\phi _g}\left( {{H_{g}^t}, y_{g}^t, \theta_g } \right)} \right) + \lambda \frac{1}{n}\sum\limits_{i = 1}^n {L\left( {{\phi _p}\left( {{P_{i}^t}, y_{i}^t, \theta_i } \right)} \right)} $$

where $L$ denotes the cross-entropy loss, $\phi_g$ is the prediction function of the model for the group activity, and $\phi_p$ is the prediction function for the actions of the individuals, $y_{g}^t$ and $y_{i}^t$ denote the labels of group activity and individual action, $n$ is number of person in the group, $\lambda$ is the weight parameter.

\subsection{Attention Module and Attentive Pooling}

As we mentioned above, persons in the scene usually have different actions, and these actions collectively define the activity of the group. In this occasion, there is a strong relativity between group activity and some person actions, while other person actions contribute less to the final group activity recognition. The aim of attentive pooling is to get a comprehensive collection of person-level information, so as to get more robust representation for the whole group and naturally change our focal object from single person to the whole group.

Inspired by success of attention mechanism applied in several tasks of computer vision and natural language processing, we derive several diversified forms of attentive pooling.

\subsubsection{Global attentive pooling (GAP)}
To reward individual actions that are clues to correctly classify a group activity, we introduce a person level weight vector $\alpha _{i}^t$ and use the vector to measure the importance of the personal actions. At time step t, given $P_{i}^t$ representing the action of ${i^{th}}$ person, firstly, we feed the personal action feature ${P_{i}^t}$ through a one-layer MLP and get $u_{i}^t$, and then compute the similarity of $u_{i}^t$ with the context vector $u_{p}^t$. Here, $u_{i}^t$ is the hidden representation of $P_{i}^t$, and $u_{p}^t$ corresponds to a person-level context vector. The aim is to get the importance of the person action by measuring the similarity of $u_{i}^t$ and $u_{p}^t$, and the weighted averages of ${P_{i}^t}$ is the feature vector of the whole group. This process can be formulated as follows:

$${u_{i}^t} = \tanh \left( {{W_{p}^t}{P_{i}^t} + {b_{p}^t}} \right)$$
$${\alpha _{i}^t} = \frac{{\exp \left( { u_{i}^t \cdot {u_{p}^t} } \right)}}{{\sum\nolimits_i {\exp \left( { u_{i}^t \cdot {u_{p}^t} } \right)} }}$$
$${G_t} = \sum\limits_{i = 1}^n {\alpha _i^tP_i^t} $$

\begin{figure}[h]
	\centering
	\includegraphics[width=8.5cm,height=3.6cm]{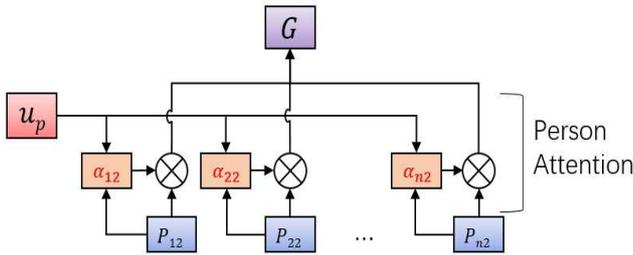}
	\caption{Global attentive pooling at time-step t}
	\label{fig3}
\end{figure}

Where $G_t$ is the feature vector representing group activity that summarizes all the information of person-level actions in the scene, $\alpha _{i}^t$ is the corresponding weight of $P_{i}^t$, $n$ is number of person in the scene. The architecture of GAP is shown in Fig. \ref{fig3}.

\subsubsection{Hierarchical attentive pooling (HAP)}

\begin{figure}[h]
	\centering
	\includegraphics[width=9.0cm,height=6.6cm]{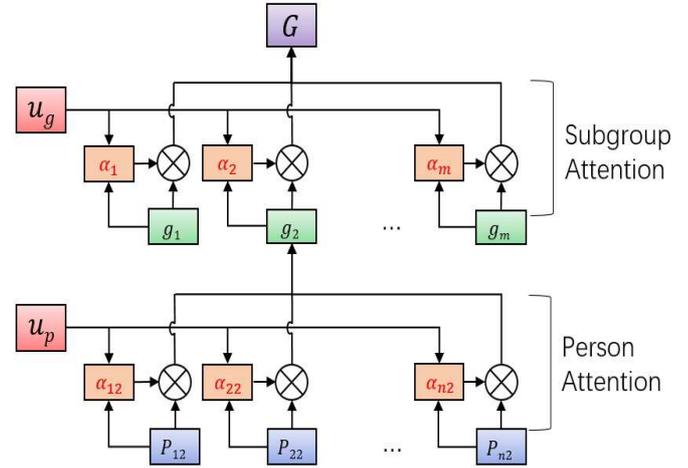}
	\caption{Hierarchical attentive pooling at time-step t}
	\label{fig4}
\end{figure}

Observing that the contributions of subgroups to the activity classification varies ,we also need to consider the subgroups discriminatingly. For instance, the global label of the whole group is "Right Winpoint", which means the volleyball team in right side of the court wins in this round. At this moment, the importance of subgroup in the right side is usually higher than that of subgroup in the left side. Inspired by the success of hierarchical attention network \cite{yang2016hierarchical}, we extend the overall person attentive pooling into hierarchical person pooling for subgroups and individuals.
First of all, we divide the group of person in the scene into several subgroups. Secondly, attentive person pooling which is similar to overall person pooling is adopted to aggregate person-level representation of each person in the subgroup, and the representations of subgroups are aggregated by attentive subgroup pooling. Finally, we get the representation of the whole group ${G_t}$, and feed it into the softmax classification layer. Here, $u_{j}^t$ is the hidden representation of $g_{j}^t$, and $u_{g}^t$ corresponds to a subgroup-level context vector. The architecture of HAP is shown in Fig. \ref{fig4}. Mathematically, for ${i^{th}}$ person in ${j^{th}}$ subgroup, person-level attentive pooling can be expressed as:

$$u_{ij}^t = \tanh \left( {W_{pj}^tP_{ij}^t + b_{pj}^t} \right)$$
$$ \alpha _{ij}^t = \frac{{\exp \left( {u{_{ij}^t} \cdot u_{pj}^t} \right)}} {{\sum _i} {\exp \left( {u{_{ij}^t} \cdot u_{pj}^t} \right)} }  $$ 
$$ g_j^t = \sum\limits_{i = 1}^{{n_j}} {\alpha _{ij}^tP_{ij}^t} $$

And subgroup-level attentive pooling is formulated as:
$$u_j^t = \tanh \left( {W_g^tg_j^t + b_j^t} \right)$$
$$\alpha _j^t = \frac{{\exp \left( {u{_j^t} \cdot u_g^t} \right)}} {{{\sum _j}\exp \left( {u{_j^t} \cdot u_g^t} \right)} }$$
$$ {G_t} = \sum\limits_{j = 1}^m {\alpha _j^tg_j^t} $$
Where $m$ is number of subgroups in the group, $n_j$ denotes number of person in ${j^{th}}$ subgroup.

\subsubsection{Integration and Separation}

As mentioned above, actions of subgroups and individuals collectively define the group activity, and the features of subgroups are aggregated in two different ways. In the previous works, some are developed in a single group, which is based on the integration of individuals or subgroups, the others handle this by considering them separately. The former are normalizing the local features (denoted as 1 group), while the latter are concatenating them (denoted as 2 groups). In order to be consistent with other approaches \citep{ibrahim2016hierarchical,shu2017cern,biswas2018structural}, we have also performed experiment where we treat the group of people as an integrated whole, and divide them into $m$ separate subgroups when utilizing the concatenation block.

\begin{figure}[h]
	\centering
	\includegraphics[width=9.2cm,height=5.5cm]{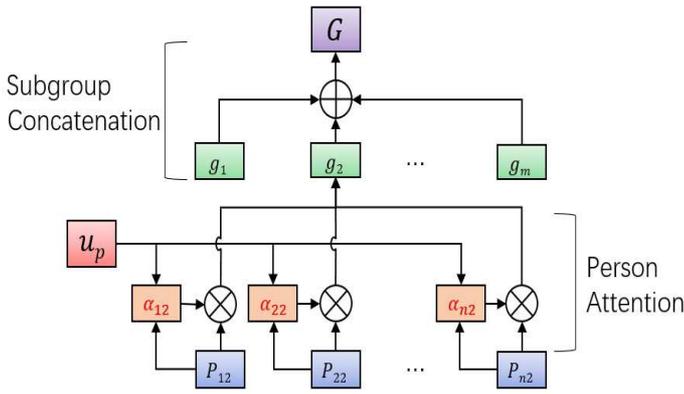}
	\caption{Attentive pooling with concatenation of separate subgroups (subgroup GAP) at time-step t}
	\label{fig5}
\end{figure}

To harness the advantage of attentive pooling and group split, we employ GAP within the subgroups and get group-level feature $G_t $ by concatenating the representations of subgroups $g_j^t$, named subgroup GAP. The architecture of subgroup GAP is shown in Fig. \ref{fig5}. Thus, $${G_t} = g_j^t \oplus g_{j + 1}^t \oplus  \cdots  \oplus g_m^t$$

In general, this method considered the whole group separately, and GAP and HAP are essentially the integration of local features.

\subsection{Implementation Details}
In accordance to \citep{ibrahim2016hierarchical}, we adopt the AlexNet \citep{krizhevsky2012imagenet} pre-trained on the ImageNet \citep{deng2009imagenet} and extract the feature vector for a person from the last convolutional layer. During training, we fine-tune the fc6 and fc7 layer. For temporal modelling, we employ LSTM implemented by Tensorflow library. The person-level LSTMs have 3000 hidden units, and group-level LSTMs have 2000 hidden units, initialized with a Gaussian distribution. And the MLP in attention module has 512 hidden units, initialized by Xavier \citep{glorot2010understanding}. For convenience, we split the whole group into 2 subgroups in the order in which the players are arranged in the annotations. We train our hierarchical networks in two steps: training the person-level networks and training the hierarchical attentive networks jointly. In the experiments, we set $\lambda=2 $, and use stochastic gradient descent with ADAM \citep{kingma2014adam}, with the initial learning rate set to 0.00001.

\section{Experiment}
\subsection{Dataset}
We evaluate our framework on the widely-used volleyball dataset \citep{ibrahim2016hierarchical}. This dataset is released in 2016, and has 55 volleyball game video documents with 4830 labelled frames, where each player is labeled and subsequently annotated with the bounding box. Each player performs one of the 9 individual actions resulting in one of the 8 group activity labels. Furthermore, the whole dataset is divided into non-overlapping sets of 24 sequences for training, 15 sequences for validation and the remaining sequences are used for testing. Similar to \citep{ibrahim2016hierarchical, shu2017cern}, we have used both training and validation sequences for training. Since not all frames are annotated by bounding boxes, the Dlib tracker \citep{king2009dlib} is used to propagate the ground-truth bounding boxes to the unannotated frames.

\subsection{Baselines}
In our experiment, we compare the proposed methods with previous baseline, and the following baselines are considered in all our experiments:
\begin{enumerate}[1. ]
	\item B1 (HTM-max pooling): This baseline model is the same as \citep{DBLP:journals/corr/IbrahimMDVM16}. We compare the performance of this method both in the experiment of integration and separation.
	\item B2 (HTM-avg pooling): This is similar to B1, but the max pooling operation has been replaced with average pooling operation when aggregating individual features. Both results in the experiment of integration and separation are shown as follows.
	\item B3 (HTM-GAP): To verify the effectiveness of overall attentive pooling, we employ GAP in the HTM. We assign the learned corresponding attention weight to each member in the group, and all the individual features would be attentively pooled. 
	\item B4 (HTM-HAP): We employ HAP in the HTM, both subgroups and individuals would be attentively pooled. Comparing to B3 model, the hierarchical attention framework is added.
	\item B5 (HTM-subgroup GAP and concatenation): In the track of separation, we employ GAP within the subgroup and concatenate the representations of subgroups in the HTM.
\end{enumerate}

\subsection{Experimental results on the volleyball dataset}

We have compared the accuracy of group activity recognition of baselines (Table \ref{table1}) and previous methods (Table \ref{table2}), and the experimental results of integration as well as separation are reported.

\begin{table}[!h]
	\processtable{Comparison of our methods with baseline methods on the volleyball dataset. \label{table1}}
	{\begin{tabular}{@{\extracolsep{\fill}}llc}\toprule
			Approach & Methods & Accuracy of group activity \\\midrule
			Integration & B1-HTM-max pooling      & 70.3\%    \\ 
			 & B2-HTM-avg pooling      & 68.5\%    \\
			 & \bfseries{B3-HTM-GAP}                       & 74.2\%    \\ 
			 & \bfseries{B4-HTM-HAP}                       & \bfseries{77.7\%}    \\  
			Separation & B1-HTM-max pooling        & 81.9\%      \\ 
		   	 & B2-HTM-avg pooling        & 80.7\%      \\
			 & \bfseries{B5-HTM-subgroup GAP}        & \bfseries{84.5\%}        \\\botrule
	\end{tabular}}{}
	
\end{table}

As shown in the Table \ref{table1}, our method using GAP or HAP module outperforms the baseline model. The results of integration show that combining the HTM and GAP module improves the accuracy of group activity, the effectiveness of proposed attentive pooling is verified. And performance is further improved by employing HAP module, which proves that attentive pooling scheme can be extended with the structural information as additional assistance. Apart from experiment of integration, the accuracy of separation is also increased by 1.5\% when comparing with B1 method. Due to the volleyball scenario where two teams distribute on the both sides of the nets, models of separation outperforms those of integration by a large margin.

\begin{table}[!h]	
	\centering
	\processtable{Comparison to the state-of-the-art on the volleyball dataset. \label{table2}}
	{\begin{tabular}{@{\extracolsep{\fill}}llcc}\toprule
			Approach	&Methods                             & Accuracy    & Year\\ \midrule
			Integration	&B1-Hierarchical LSTM  		\citep{DBLP:journals/corr/IbrahimMDVM16}      & 70.3\% 	 & 2016 \\ 
			&CERN (1group) \citep{shu2017cern}                       			& 73.5\% 	 & 2017 \\
			&SBGAR \citep{li2017sbgar}											& 66.9\% 	 & 2017	\\
			&SRNN (1group) \citep{biswas2018structural}              			& 74.4\%	 & 2018	 \\ 
			&\bfseries{Ours-B3}            & \bfseries{77.7\%}         & \\ 
			Separation		&B1-Hierarchical LSTM ) 	\citep{DBLP:journals/corr/IbrahimMDVM16}     & 81.9\%  & 2016\\ 
			&CERN (2 groups) \citep{shu2017cern}                     & 83.3\%  & 2017 \\ 
			&SRNN (2 groups) \citep{biswas2018structural}                     & 83.4\%  & 2018 \\ 
			&\bfseries{Ours-B4}      & \bfseries{84.5\%}   & \\         \botrule
	\end{tabular}}{}
	
\end{table}

Table \ref{table2} compares the proposed methods with previous methods on this dataset. In experiment of integration, our HAP based model achieves higher accuracy than previous methods, and the performance of subgroup GAP based model is better than CERN and baseline method and comparable to the SRNN model in experiment of separation. All the experimental results are achieved based on the Alexnet features.

\subsection{Discussion}
In this part, we firstly discuss the impact of hidden units in attention modules and the confusion matrix of prediction. Then we list some of the experiment results, and analyse how our proposed methods works. 

\begin{table}[!h]
	\processtable{Comparison of different numbers of MLP hidden units in attention module. \label{table3}}
	{\begin{tabular}{@{\extracolsep{\fill}}clc}\toprule
			Numbers of hidden units  & Methods & Accuracy of group activity \\\midrule
			1024 & HTM-GAP                       & 69.0\%    \\ 
				 & HTM-HAP                       & \bfseries{77.7\%}    \\  
				 & HTM-subgroup GAP			  & 83.1\%	  \\	
		    512  & HTM-GAP                       & \bfseries{74.2\%}    \\ 
		   		 & HTM-HAP                       & 77.4\%    \\  
		  	 	 & HTM-subgroup GAP			  & \bfseries{83.4\%}	 \\  
    	 	256  & HTM-GAP                       & 73.3\%    \\ 
		  	 	 & HTM-HAP                       & 77.4\%    \\  
		  	 	 & HTM-subgroup GAP			  & 82.4\%	  \\ \botrule
	\end{tabular}}{}

\end{table}
To evaluate the impact of MLP hidden units in attention module, we conduct a set of experiments outlined Table \ref{table3}. Both in experiment of GAP and subgroup GAP, we get better performance when employing 512 hidden units. And the accuracy are higher when 1024 hidden units are used in experiment of HAP.   

\begin{figure*}[htb]
	\centering
	\subcaptionbox{Original frame in the dataset\label{subfig:left}}[.32\linewidth]
	{
		\includegraphics[width= .32\linewidth]{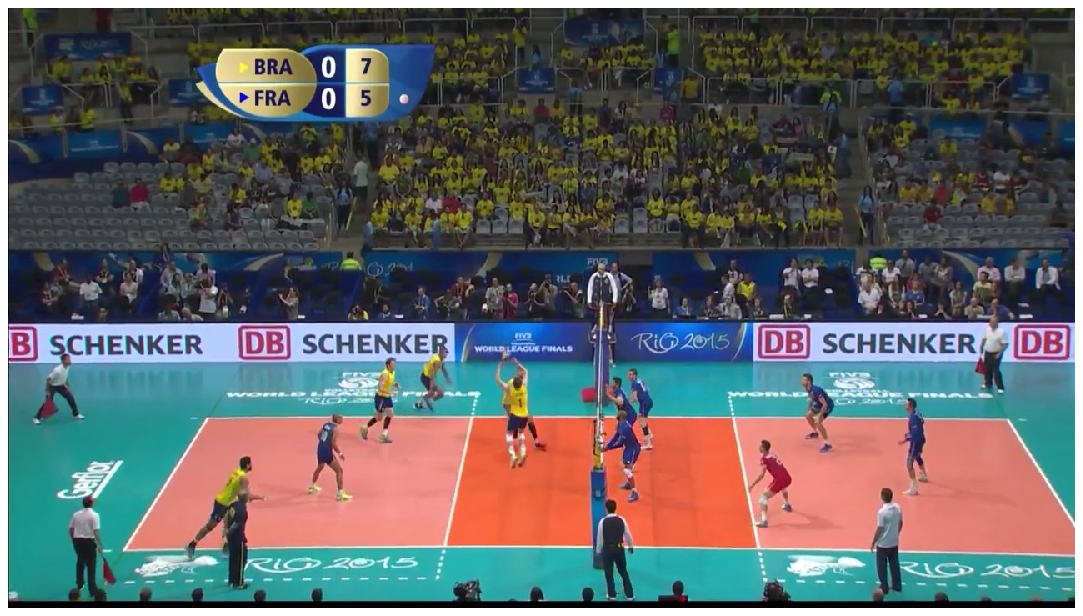} 
		\includegraphics[width= .32\linewidth]{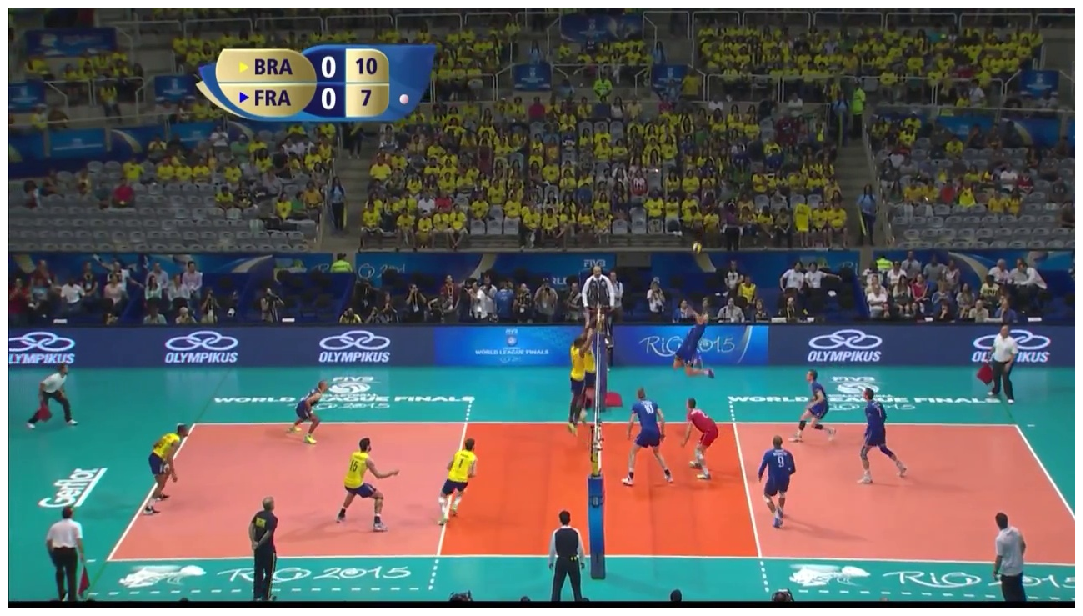}
		\includegraphics[width= .32\linewidth]{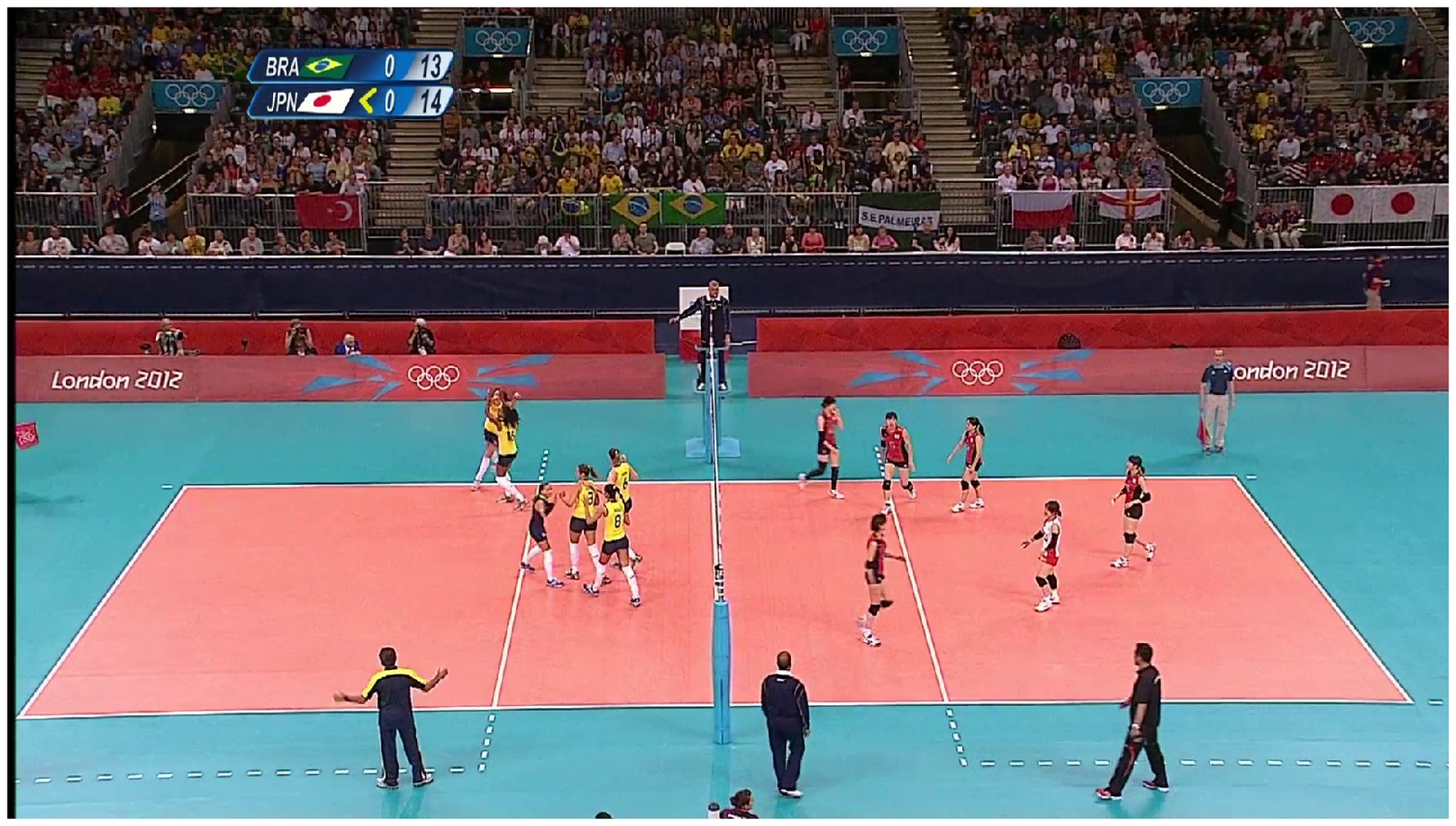}}\quad
	\subcaptionbox{Frame with individual attention weights assigned\label{subfig:middle}}[ .32\linewidth]
	{
		\includegraphics[width= .32\linewidth]{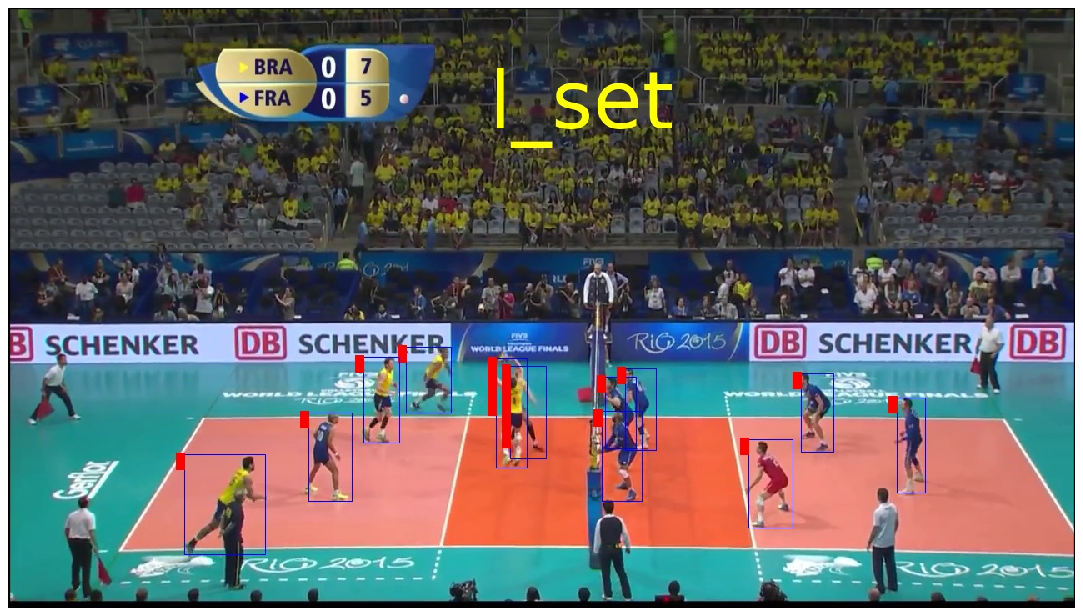}
		\includegraphics[width= .32\linewidth]{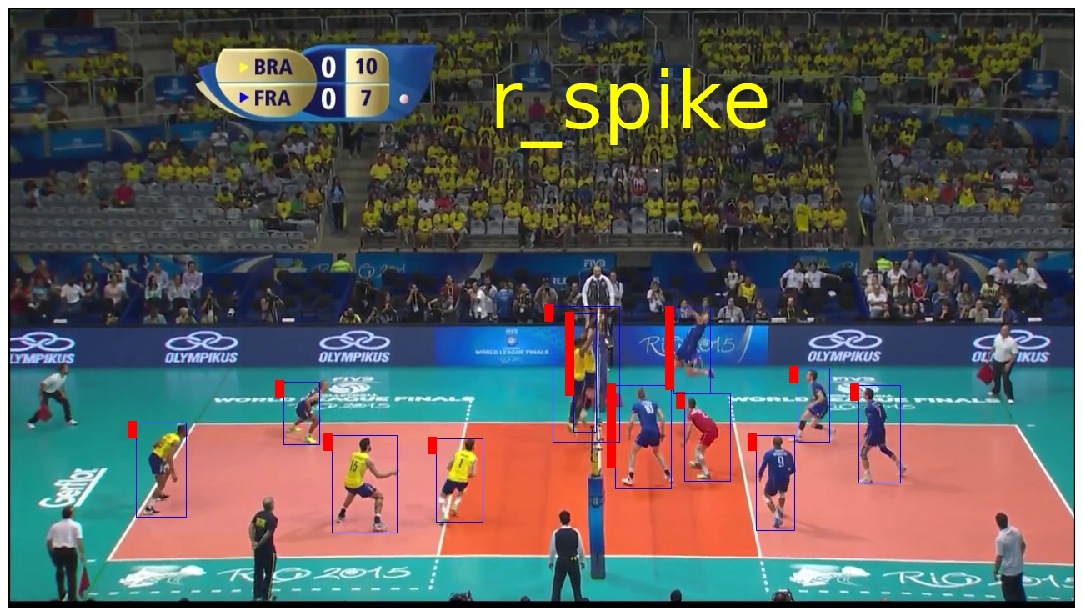}
		\includegraphics[width= .32\linewidth]{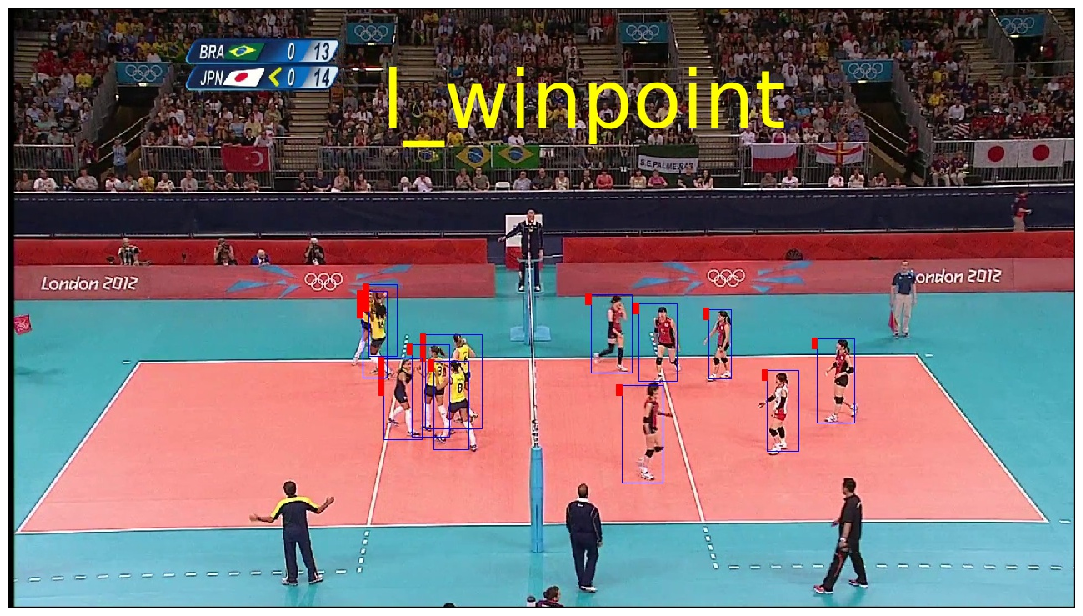}}\quad
	\subcaptionbox{Frame with individual spotlight according to the attention weights\label{subfig:right}}[ .32\linewidth]
	{
		\includegraphics[width= .32\linewidth]{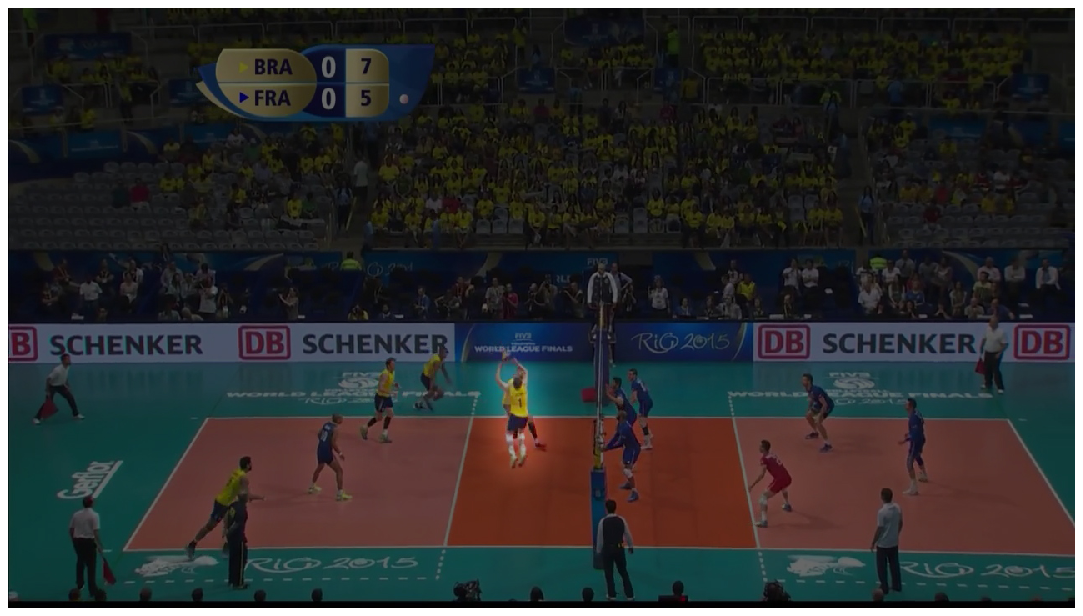}
		\includegraphics[width= .32\linewidth]{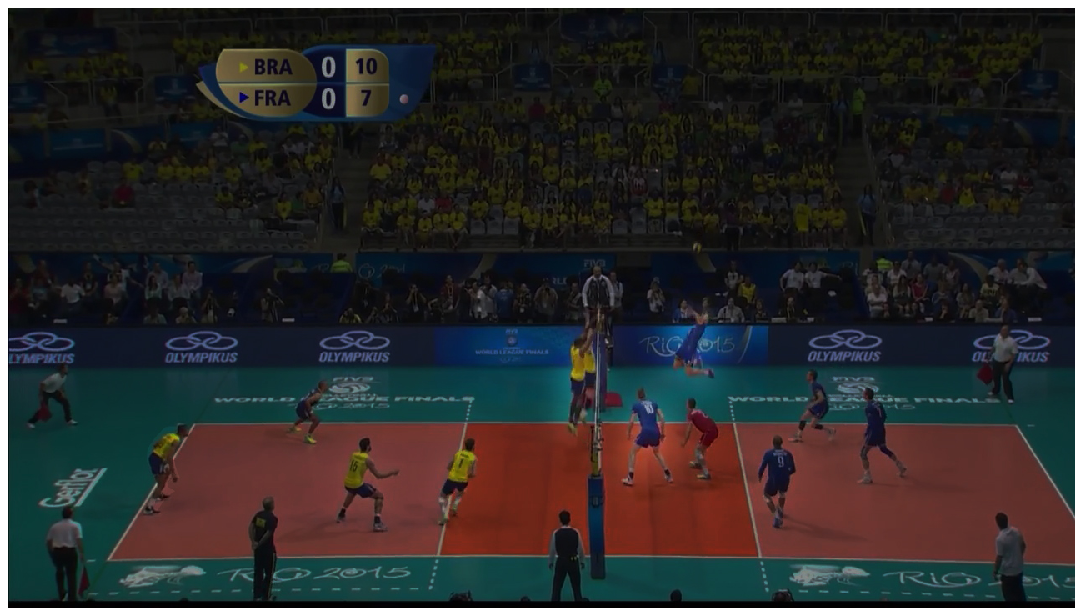}
		\includegraphics[width= .32\linewidth]{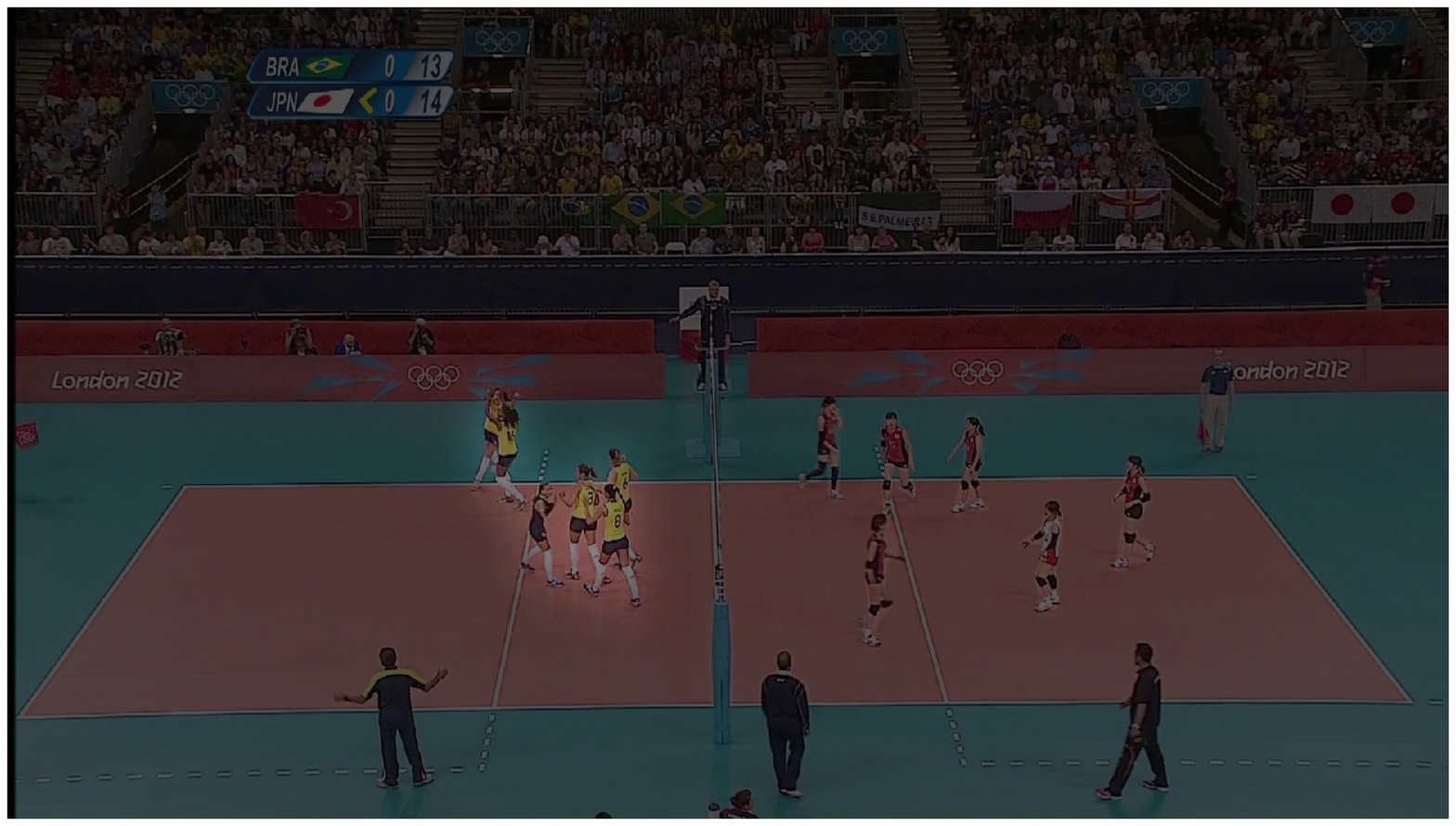}}

	\caption{Examples of visual results (better viewed in colour). In figure (b), the length of bold red lines annotated alongside with bounding boxes corresponds to the value of attention weights. The longer the red line is, the more important it shows. In figure (c), we highlight the important individuals according to the value of attention weights. Examples shows that the attentive pooling method can not only improve the accuracy of group activity, but makes the model more interpretable. \label{fig:{original images from dataset}_{images with individual attention weights assigned}_{images with individual spotlight}}}
	
\end{figure*}

\begin{figure}[h]
	\centering
	\includegraphics[width=6.8cm,height=6.5cm]{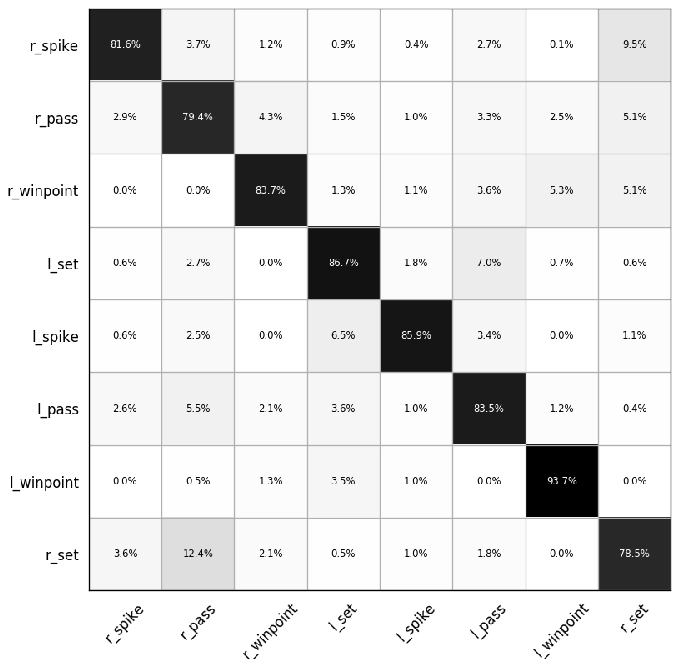}
	\caption{Confusion matrix on the Volleyball Dataset obtained by using our attentive pooling based model.}
\end{figure}

Figures 6 and 7 show the attention map of group activities using our model and the visualizations of the confusion matrix. The case study shown in Figure 6 illustrates how attentive pooling works, we choose to give three types of group activities as typical examples. In the 1st row, the frame is labelled as "Left Set". The longer red line around the player who is setting the ball for his teammates indicates the higher attention weights he has. And it is easier to figure out the different contribution of each person when we highlight the important individuals. Similarly, the frame labelled as "Right Spike" shown in the 2nd row highlights  the players who perform spiking and blocking in the court. Intuitively, we can recognize the group activity just by considering the highlighted individual actions. The bottom frame which is labelled as "Left Winpoint" is slight different from frames above. In this frame, the volleyball players in the left side of the court are gathering, while their rivals are scattered on the other side of the court. The results show that the average attention weights of players who gathered round are higher than that of scattered players, which makes it easier to recognize the group label "Left Winpoint". The learned attention weights not only strengthen the positive effect of difference on recognition, but cooperate well with the unity of individuals. 

From the confusion matrix shown in Figure 7, we can observe that "Left Winpoint" gains highest accuracy, while the accuracy of "Right Set" and "Right pass" are lower than 80\%. The aggregation of players could be easily recognized based our model, which makes the high performance of winpoint activities. Due to the complexity of volleyball scenario and the similarity of set and pass activities, there is some confusion when distinguishing them. Besides, the interval between setting and spiking is always very short, which makes it difficult to recognize.

\section{Conclusions}
In this work, we address the motivation of taking the different contribution each part of group makes into consideration with a novel pooling scheme based on attention mechanism, named attentive pooling. The GAP attentively selects the focal parts inside the group, and HAP extends the GAP by modelling structure of the group in a hierarchical manner. By utilizing those variants of attentive pooling, the proposed model can generate more discriminative and interpretable representations of the group activity. We evaluated the models on the widely-used Volleyball Dataset, and experimental results showed the superiority of the proposed model. 

\section{Acknowledgements}
The authors are thankful for the financial support from the National Natural Science Foundation of China (U1636220, 61432008, 61472423).

{\small
	\bibliographystyle{iet}
	\bibliography{CTAbib}

\begin{thebibliography}{10}

\bibitem{simonyan2014two}
Simonyan, K., Zisserman, A.: `Two-stream convolutional networks for action
  recognition in videos'.
\newblock Advances in neural information processing systems (NIPS 2014),  2014.
  pp.~ 568--576

\bibitem{wang2016temporal}
Wang, L., Xiong, Y., Wang, Z., \emph{et~al.}: `Temporal segment networks:
  Towards good practices for deep action recognition'.
\newblock 2016 European Conf. on Computer Vision (ECCV),  2016. pp.~ 20--36

\bibitem{ji20133d}
Ji, S., Xu, W., Yang, M., Yu, K.: `3d convolutional neural networks for human
  action recognition', \emph{IEEE transactions on pattern analysis and machine
  intelligence},  2013, \textbf{35}, (1), pp.~221--231

\bibitem{tran2015learning}
Tran, D., Bourdev, L., Fergus, R., Torresani, L., Paluri, M.: `Learning
  spatiotemporal features with 3d convolutional networks'.
\newblock 2015 IEEE Conf. on Computer Vision and Pattern Recognition (CVPR),
  2015. pp.~ 4489--4497

\bibitem{yu2017fully}
Yu, S., Cheng, Y., Xie, L., Li, S.Z.: `Fully convolutional networks for action
  recognition', \emph{IET Computer Vision},  2017, \textbf{11}, (8),
  pp.~744--749

\bibitem{varol2018long}
Varol, G., Laptev, I., Schmid, C.: `Long-term temporal convolutions for action
  recognition', \emph{IEEE transactions on pattern analysis and machine
  intelligence},  2018, \textbf{40}, (6), pp.~1510--1517

\bibitem{choi2012unified}
Choi, W., Savarese, S.: `A unified framework for multi-target tracking and
  collective activity recognition'.
\newblock 2012 European Conf. on Computer Vision (ECCV),  2012. pp.~ 215--230

\bibitem{ibrahim2016hierarchical}
Ibrahim, M.S., Muralidharan, S., Deng, Z., Vahdat, A., Mori, G.: `A
  hierarchical deep temporal model for group activity recognition'.
\newblock 2016 IEEE Conf. on Computer Vision and Pattern Recognition (CVPR),
  2016. pp.~ 1971--1980

\bibitem{shu2017cern}
Shu, T., Todorovic, S., Zhu, S.C.: `Cern: confidence-energy recurrent network
  for group activity recognition'.
\newblock 2017 IEEE Conf. on Computer Vision and Pattern Recognition (CVPR),
  2017. pp.~ 5523--5531

\bibitem{li2017sbgar}
Li, X., Chuah, M.C.: `Sbgar: Semantics based group activity recognition'.
\newblock 2017 IEEE Conf. on Computer Vision and Pattern Recognition (CVPR),
  2017. pp.~ 2876--2885

\bibitem{bagautdinov2017social}
Bagautdinov, T.M., Alahi, A., Fleuret, F., Fua, P., Savarese, S.: `Social scene
  understanding: End-to-end multi-person action localization and collective
  activity recognition.'.
\newblock 2017 IEEE Conf. on Computer Vision and Pattern Recognition (CVPR),
  2017. pp.~ 3425--3434

\bibitem{wang2017recurrent}
Wang, M., Ni, B., Yang, X.: `Recurrent modeling of interaction context for
  collective activity recognition'.
\newblock 2017 IEEE Conf. on Computer Vision and Pattern Recognition (CVPR),
  2017. pp.~ 3048--3056

\bibitem{biswas2018structural}
Biswas, S., Gall, J.: `Structural recurrent neural network (srnn) for group
  activity analysis'.
\newblock 2016 IEEE Winter Conf. on Applications of Computer Vision (WACV),
  2018. pp.~ 1625--1632

\bibitem{DBLP:journals/corr/IbrahimMDVM16}
Ibrahim, M.S., Muralidharan, S., Deng, Z., Vahdat, A., Mori, G.: `Hierarchical
  deep temporal models for group activity recognition', \emph{arXiv preprint
  arXiv:1607.02643},  2016,

\bibitem{LSTM}
Hochreiter, S., Schmidhuber, J.: `Long short-term memory', \emph{Neural
  computation},  1997, \textbf{9}, (8), pp.~1735–1780

\bibitem{herath2017going}
Herath, S., Harandi, M., Porikli, F.: `Going deeper into action recognition: A
  survey', \emph{Image and vision computing},  2017, \textbf{60}, pp.~4--21

\bibitem{chaquet2013survey}
Chaquet, J.M., Carmona, E.J., Fern{\'a}ndez.Caballero, A.: `A survey of video
  datasets for human action and activity recognition', \emph{Computer Vision
  and Image Understanding},  2013, \textbf{117}, (6), pp.~633--659

\bibitem{nabi2013temporal}
Nabi, M., Bue, A., Murino, V.: `Temporal poselets for collective activity
  detection and recognition'.
\newblock 2013 IEEE Conf. on Computer Vision Workshops (ICCV Workshops),  2013.
  pp.~ 500--507

\bibitem{vaswani2017attention}
Vaswani, A., Shazeer, N., Parmar, N., \emph{et~al.}: `Attention is all you
  need'.
\newblock Advances in neural information processing systems (NIPS 2017),  2017.
  pp.~ 5998--6008

\bibitem{mnih2014recurrent}
Mnih, V., Heess, N., Graves, A.: `Recurrent models of visual attention'.
\newblock Advances in neural information processing systems (NIPS 2014),  2014.
  pp.~ 2204--2212

\bibitem{sharma2015action}
Sharma, S., Kiros, R., Salakhutdinov, R.: `Action recognition using visual
  attention', \emph{arXiv preprint arXiv:1511.04119},  2015,

\bibitem{yang2016hierarchical}
Yang, Z., Yang, D., Dyer, C., He, X., Smola, A., Hovy, E.: `Hierarchical
  attention networks for document classification'.
\newblock 2016 Conference of the North American Chapter of the Association for
  Computational Linguistics: Human Language Technologies,  2016. pp.~
  1480--1489

\bibitem{videolstm}
Li, Z., Gavves, E., Jain, M., Snoek, C.G.M.: `Videolstm convolves, attends and
  flows for action recognition', \emph{arXiv preprint arXiv:1607.01794},  2016,

\bibitem{attentionclusters2019}
Long, X., Gan, C., de~Melo, G., Wu, J., Liu, X., Wen, S.: `Attention clusters:
  Purely attention based local feature integration for video classification'.
\newblock 2018 IEEE Conf. on Computer Vision and Pattern Recognition (CVPR),
  2018. pp.~ 7834--7843

\bibitem{donahue2015long}
Donahue, J., Anne.Hendricks, L., Guadarrama, S., Rohrbach, M., Venugopalan, S.,
  Saenko, K., \emph{et~al.}: `Long-term recurrent convolutional networks for
  visual recognition and description'.
\newblock 2015 IEEE Conf. on Computer Vision and Pattern Recognition (CVPR),
  2015. pp.~ 2625--2634

\bibitem{krizhevsky2012imagenet}
Krizhevsky, A., Sutskever, I., Hinton, G.E.: `Imagenet classification with deep
  convolutional neural networks'.
\newblock Advances in neural information processing systems (NIPS 2012),  2012.
  pp.~ 1097--1105

\bibitem{deng2009imagenet}
Deng, J., Dong, W., Socher, R., Li, L.J., Li, K., Fei.Fei, L.: `Imagenet: A
  large-scale hierarchical image database'.
\newblock 2009 IEEE Conf. on Computer Vision and Pattern Recognition (CVPR),
  2009. pp.~ 248--255

\bibitem{glorot2010understanding}
Glorot, X., Bengio, Y.: `Understanding the difficulty of training deep
  feedforward neural networks'.
\newblock Proc. of the 13th international conference on artificial intelligence
  and statistics (AISTATS-2010),  2010. pp.~ 249--256

\bibitem{kingma2014adam}
Kingma, D.P., Ba, J.: `Adam: A method for stochastic optimization', \emph{arXiv
  preprint arXiv:1412.6980},  2014,

\bibitem{king2009dlib}
King, D.E.: `Dlib-ml: A machine learning toolkit', \emph{Journal of Machine
  Learning Research},  2009, \textbf{10}, (Jul), pp.~1755--1758

\end{thebibliography}
}

\end{document}